# Design of an innovative robotic surgical instrument for circular stapling

Paul Tucan[1][0000-0001-5660-8259], Nadim Al Hajjar[2][0000-0001-5986-1233], Calin Vaida [1][0000-0003-2822-9790], Alexandru Pusca [1][0000-0002-5804-575X], Corina Radu[3][0000-0003-0005-0262], Daniela Jucan[1][0009-0004-0219-9858], Tiberiu Antal[1][0000-0002-0042-5258], Adrian Pisla [1][0000-0002-5531-6913], Damien Chablat[1,4][0000-0001-7847-6162], and Doina Pisla[1,5, *][0000-0001-7014-9431]

[1] CESTER, Research Center for Industrial Robots Simulation and Testing, Technical University of Cluj-Napoca, 28 Memorandumului Street, 400114 Cluj-Napoca, Romania
[2] Department of General Surgery, University of Medicine and Pharmacy "Iuliu Hațieganu", 400347 Cluj-Napoca, Romania
[3] Department of Internal Medicine, "Iuliu Hatieganu" University of Medicine and Pharmacy, 400347 Cluj-Napoca, Romania
[4] École Centrale Nantes, Nantes Université, CNRS, LS2N, UMR 6004, F-44000 Nantes, France
[5] Technical Sciences Academy of Romania, 26 Dacia Bvd, 030167 Bucharest, Romania
*Corresponding author: doina.pisla@mep.utcluj.ro

**Abstract.** Esophageal cancer remains a highly aggressive malignancy with low survival rates, requiring advanced surgical interventions like esophagectomy. Traditional manual techniques, including circular staplers, face challenges such as limited precision, prolonged recovery times, and complications like leaks and tissue misalignment. This paper presents a novel robotic circular stapler designed to enhance the dexterity in confined spaces, improve tissue alignment, and reduce post-operative risks. Integrated with a cognitive robot that serves as a surgeon's assistant, the surgical stapler uses three actuators to perform anvil motion, cutter/stapler motion and allows a 75-degree bending of the cartridge (distal tip). Kinematic analysis is used to compute the stapler tip's position, ensuring synchronization with a robotic system.

## 1 Introduction

Esophageal cancer is one of the most malignant forms of cancer worldwide due to its aggressive nature and reduced rate of survival, with a five-year survival rate around 15% to 25 % [1]. Some of the treatment options for esophageal cancer include surgery (esophagectomy) [2-4], radiation therapy (external beam radiation or brachytherapy) [5-8], chemotherapy [2-4], targeted therapy [9], immunotherapy [2,9], endoscopic treatment (mucosal resection and radiofrequency ablation) [6] and palliative care [5-8]. The treatment option for esophageal cancer depends on multiple factors, among them, cancer type (adenocarcinoma-mostly linked to the obesity and chronic acid reflux or squamous cell carcinoma-associated with smoking and alcohol), stage of the cancer, location and overall health state of the patient. Esophagectomy is the gold standard for esophageal cancer as long as the cancer is detected in early stage and the patient is fit for the surgery [10]. Manual esophagectomy can be performed either as open surgery or minimally invasive surgery and is often associated with limited precision in confined spaces, prolonged recovery times and higher risk of complications (infections or nerve damage). Using robotic systems or automatic devices to perform surgery may address the previously described drawbacks of the manual surgery by providing enhanced dexterity using articulated instruments, superior 3D visualization and minimally invasive techniques that may reduce the tissue trauma, improve the patient outcomes and reduce recovery times. Over time several robotic systems for anastomosis were developed. STAR (Smart Tissue Autonomous Robot) [11] was developed for soft tissue anastomosis showing promising results in esophageal applications. The system uses advanced computer vision and machine learning to autonomously perform sutures with high accuracy, minimizing post operatory complications (leaks), the drawbacks of the system include high computational complexity and cost. Milone et al [12] evaluates robotic hand-sewn anastomosis in Ivor Lewis esophagectomy, emphasizing precision in confined spaces. The technique demonstrates comparable leak rates to stapling methods, suggesting it as a viable alternative. However, increased operative times and technical expertise requirements are noted. Multicenter trials are recommended to refine its application. Guerra et al [13] explores the feasibility of side-to-side linear-stapled anastomosis using robotic systems. Benefits include shorter operative times and uniform staple lines, potentially reducing complications. The study highlights technical challenges with device compatibility and surgeon proficiency. Findings encourage further investigation into efficiency and safety outcomes. Wee et al [14] evaluates the circular stapler for its ability to standardize the procedure and reduce operative complexity. Early findings suggest favorable outcomes, but challenges with stapler positioning and device limitations are noted. Further research is necessary to establish long-term clinical benefits and operational feasibility.

This paper presents the design of an innovative robotic circular stapler [15] used for anastomosis in esophageal cancer surgery. The newly designed instrument maintains the advantages of the classical surgical stapler, providing increased dexterity in confined spaced (stomach-esophagus volume) and by automation, increasing the procedure accuracy and reproducibility, reducing post operatory risks like leaks, tissue misalignments or infection. The stapler is integrated with a cognitive robot able of running artificial intelligence agents, collaborative tasks and voice control making it suitable for further implementation as surgeon assistant. The circular stapler has one Degree of Freedom allowing the bending of its distal tip after insertion, making it suitable for use in minimally invasive esophagogastric anastomosis. Moreover, its integration with a cognitive robotic system allows access to artificial intelligence agents able to be implemented within the robot that could provide real time 3D navigation during the surgery.

The paper is structured as follows: Section 2 presents several data regarding the esophagectomy and esophagogastric anastomosis; Section 3 presents a SWOT analysis of circular staples used in conventional anastomosis and several interventions are pointed out to enhance the capabilities of the staplers and improve the surgery outcomes; Section 4 presents the design of the circular stapler followed by its integration with a cognitive robot in section 5, Conclusions and Acknowledgments.

## 2      Esophagectomy and anastomosis

Medical protocol for esophagectomy [16] includes several instruments required for the surgical procedure, depending on the medical approach chosen (trans hiatal [17], Ivor Lewis [18], McKeown [19] or thoracoabdominal [20]). This paper analyses the medical protocol for Ivor Lewis approach to design the circular stapler. In the case of Ivor-Lewis esophagectomy, the circular stapler is used for esophagogastric anastomosis (a surgical connection between the remaining part of the esophagus and the stomach after tumor resection). Fig.1 presents the esophagogastric anastomosis. During the esophagectomy, the part of esophagus containing the tumor is removed and, in order to allow the digestive system to perform the digestion function the remaining part of the esophagus is pulled down while the upper part of the stomach is pulled-up and connected with the esophagus using the circular stapler. The anastomosis is performed using a circular stapler (Fig.2). The anvil of the circular stapler can be detached from the stapler and inserted into the remaining part of the esophagus after resection, and, using manual suture the esophageal walls are sutured around the anvil shaft.

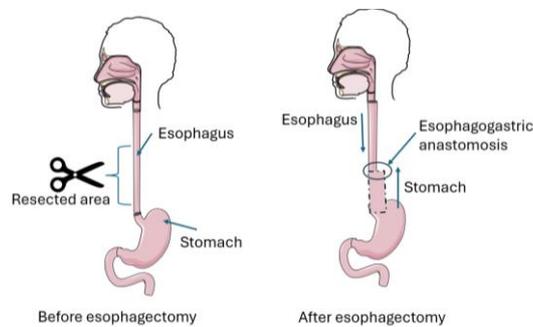

**Fig.1.** Esophagogastric anastomosis [21]

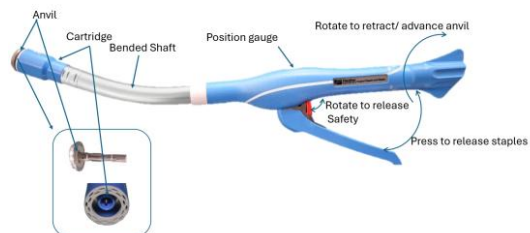

**Fig. 2**. Circular stapler [22]

After the suture, the stapler with the anvil connecting tip retracted is inserted into the stomach, and by advancing the anvil connecting tip, the stomach wall is penetrated. The anvil connecting tip is advanced to its maximum position and using a laparoscopic instrument the anvil is connected to connecting tip. After the connection the anvil is retracted, and the two organs are joined together. The anvil is retracted until the position gauge indicates the proper position for stapling procedure. Before pressing the stapling handle, 10-20 seconds delay is required in order for the tissue of esophagus to properly connect to the stomach. After the delay the handle safety is deactivated and using steady and constant pressure the stapling handle is actuated. By pressing the stapling handle, the staples from the cartridge are ejected and after the stapling, a circular knife cuts the tissue inside the stapling circle. After retracting the stapler, this should contain two tissue "donuts" from the esophageal and stomach wall. In order for the procedure to be successful these pieces of tissue should have a circular form, with no interruptions, any deviation from this geometry may result in anastomosis leaking.

## 3     SWOT analysis of the surgical circular staplers

In order to determine the effectiveness of the circular staplers in esophagectomy, a SWOT(Strengths-Weaknesses-Opportunities-Threats) analysis was conducted (Fig. 3) based on medical personal expertise involved in the research project.

**Strengths**
- ✓ **Efficiency** (Significantly reduces the time required for creating anastomoses compared to manual suturing)
- ✓ **Precision** (Creates uniform, consistent staples that improve surgical outcomes.)
- ✓ **Minimally Invasive** (Facilitates minimally invasive surgeries, such as laparoscopic and robotic-assisted procedures.)
- ✓ **Broad Application** (Used in various surgical fields, including gastrointestinal, colorectal, and bariatric surgeries)
- ✓ **Proven Effectiveness** (High success rates in achieving watertight anastomoses and reducing leak rates)

**Weaknesses**
- ✓ **Cost** (High device cost and recurring expense due to single use)
- ✓ **Technical Challenges** (Requires precise placement and alignment)
- ✓ **Size Limitation** (Fixed diameters may not suit all anatomical conditions or patient needs)
- ✓ **Postoperative Risks** (Potential for anastomotic strictures, leaks, or tissue necrosis.)
- ✓ **Ergonomics** (The use of the instrument is restricted by the adjacent organs)

**Opportunities**
- ✓ **Technological Advancements** (Development of more flexible, smart, or customizable staplers to address size and alignment issues)
- ✓ **Market Expansion** (Increasing demand for minimally invasive procedures in emerging markets)
- ✓ **Enhanced Safety Features** (Integration of real-time feedback or sensors to minimize complications)
- ✓ **Applications in New Procedures** (Potential use in novel surgeries, such as organ transplantation or advanced bariatrics)

**Threats**
- ✓ **Competition** (Emerging technologies, such as tissue adhesives or robotic suturing systems, may reduce reliance on staplers)
- ✓ **Regulatory Challenges** (Stricter safety and quality standards could increase costs and delay innovation)
- ✓ **Economic Pressure** (High costs may limit adoption in resource-constrained healthcare systems)
- ✓ **Patient Specific Complications** (Anatomical variability or comorbidities could increase complication risks.)

**Fig. 3.** SWOT analysis of surgical circular staplers

Following the SWOT analysis and several meetings with surgeons using surgical staplers in esophagogastric anastomosis, several characteristics for a robotic assisted stapler resulted. Table 1 presents three main intervention areas identified: flexibility, ergonomics and automation.

**Table 1.** Intervention areas for innovative design of robotic circular stapler

| Intervention area | Characteristic | Description |
| --- | --- | --- |
| Flexibility | Tissue Adaptability | Develop dynamic tissue sensing technology to detect and adjust compression and stapling force |
| | Multi-Size Compatibility | Implement flexible stapling heads with multi-angle articulation to adapt to various surgical sites and anatomical constraints. |
| Ergonomics | Lightweight Design | Use lightweight, biocompatible materials like high-grade polymers to reduce the weight of the instrument. |
| | Robotic integration | Develop ergonomic controls for robotic integration |
| Automation | Smart Stapling control | Integrate AI algorithms for predictive stapling force and alignment based on preoperative imaging data |
| | Cartridge Replacement | Develop interchangeable cartridges for the instrument, possible automatic cartridge changing device |
| | Error Detection and Correction | Use of Machine Learning to identify potential alignment or pressure issues during the procedure and automatically adjust parameters to correct them. |
| | Real-Time Monitoring and Feedback | Use sensors to monitor tissue response (elasticity, tension) and provide data to the surgeon. |

Table 1 can represent the starting point in defining the characteristics of the new robotic assisted circular stapler and represents a clear and strategic direction for future advancements.

## 4    Design of the robotic circular stapler

With respect to the previous section, the robotic circular stapler was designed (Fig. 4). The anvil-stapler-cutting ensemble is similar with the classical manual instrument, the robotic instrument is able to bend to a maximum of 75 degrees, and for the actuation 3 motors are used. The instrument can be attached to the robot through a flange embedding DIN ISO 9409-1-50-7-M6 hole pattern.

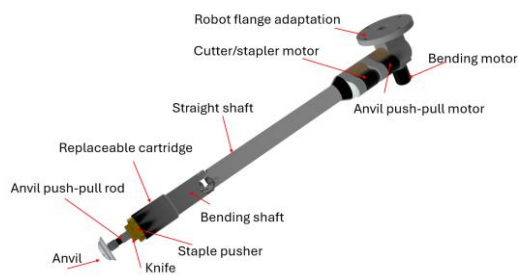

**Fig. 4.** Design of the robotic circular stapler

The robotic circular stapler uses three motors to actuate the mechanisms required for anvil motion, knife motion and the bending of the instrument. Fig. 5 describes the anvil motion mechanism. The head of the anvil is detachable from the instrument as long as the push-pull mechanism is at its full extended range, when the anvil is retracted it cannot be detached anymore, to eliminate the risk of detaching while stapling. The detached anvil is used to compress the tissue on the surface of the cartridge and to assure equal compression in order to avoid misalignment that can cause anastomosis leakages. The mechanism is actuated by a stepper motor driving a helical gearbox. The advancement and retraction of the anvil is made using a rod with a trapezoidal screw, advanced with the help of a nut embedded into the helical gearbox. The pressure applied by the anvil head upon the tissue is very important. A study published by Shiraishi et al [23] in 2022 showed that applying little to no tension reduces the anastomotic stricture (the narrowing of the surgical anastomosis caused by excessive scar tissue formation, reduced blood flow or other healing complications) with almost 4%. For real-time monitoring of the pressure applied on the tissue, force sensors are mounted on the anvil head. The compression force for esophagogastric anastomosis is generally in the range of 2.5 -5 $N/mm^2$. Using this pressure, the required motor torque for actuating the anvil mechanism is determined using Eq.1

$$T = \frac{F \cdot d_m \cdot (\tan(\phi) + \mu)}{2 \cdot \pi} \tag{1}$$

where, T is the torque (Nm), F is the axial load (N), $d_m$ is the mean diameter of the thread (0.0085 m), $\phi$ is the lead angle of the thread (0.112 radians), and $\mu$ =0.15 is the coefficient of friction between the nut and the screw. The axial load F=392.7 N and is obtained using formula F=P·A where P is the pressure ($N/mm^2$) and A is the cross-sectional area of the surface in contact. The resulting torque for the motor is T=0.139 Nm.

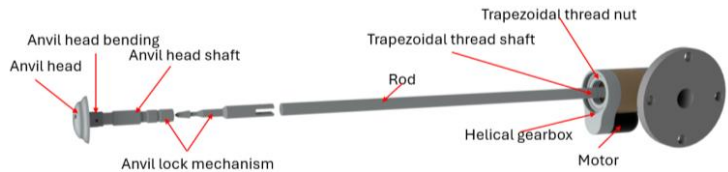

**Fig.5.** Anvil mechanism

The cutting and stapling mechanism is activated after the anvil has reached the required contact pressure. The mechanism is actuated by a stepper motor through a helical gearbox that contains an M16 threaded nut used for advancing and retracting a M16 threaded shaft connected to the cutting/stappling head. When the motor for this mechanism is actuated, the straight rod is pushed forward and presses the staples pusher and the circular knife. The staples are ejected from the cartridge and staple the tissue creating stomach-esophagus bond, at the same time the circular knife is ejected using the same mechanism and the tissue of the esophagus and the stomach is cut into a circular form ("donut") creating the anastomosis. After the anastomosis, the contact pressure is maintained for several seconds before retracting the cutting/stapling mechanism. Using the same equation (Eq.1) the required torque for the cutting/stapling motor is 0.02 Nm.

The bending of the distal head is performed using a cable mechanism. The working principle and maximum bending are presented in Fig.7. The torque required for the bending of the distal head is determined using Eq.2

$$T = F \cdot r \qquad (2)$$

where, F is the force exerted by the motor on the shaft, distributed to the cables, and r is the radius at which the cables are attached to the shaft. By imposing a 200 N force, r =15 mm and including the 1:30 worm gear box ratio a total required torque of 0.13 Nm is obtained for the motor.

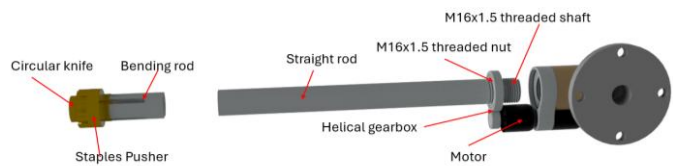

**Fig.6.** Cutter /stapler mechanism

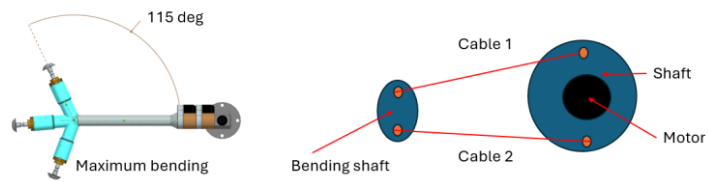

**Fig.7.** Bending mechanism

## 5    Robot integration

The circular stapler is integrated into the surgical procedure with the help of a robot. Fig.8a presents the configuration of the system during the stapling procedure, while Fig. 8b presents the bending of the instrument inside the patient after the insertion. The robotic system contains a cognitive robot (MAiRA Pro M [24]) able to perform collaborative tasks, integrate AI algorithms and voice control commands, making it suitable for further development as a surgery assistant. The circular stapler is attached to the flange of MAiRA robot, and the robot is controlled during the surgical procedure using voice commands. The circular stapler is inserted inside the patient between the 5$^{th}$ and 6$^{th}$ ribs and then is bent to follow the esophagus path in order to perform the anastomosis. During the procedure, the robot is controlled using its own controlling software to insert and position the instrument (using voice control), after the insertion of the instrument, the bending, anvil motion, and stapling operations are performed using the control system of the robotic circular stapler.

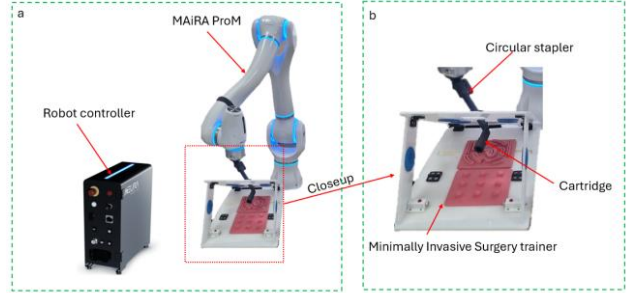

Fig. 8. The robotic configuration using the new circular stapler: a. robot configuration during the surgery; b. bending of the instrument during the surgery.

The kinematic scheme for determining the coordinates from the flange of the robot to the tip of the instrument is presented in Fig. 9.

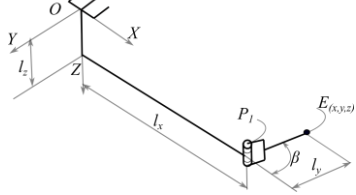

**Fig.9** Kinematic scheme for computing coordinates at the tip of the stapler

The parameters required to determine the coordinates at the tip of the stapler (point E) are $l_x$ (the distance along the X axis from the origin O to pivot point $P_1$), $l_y$ (the length of the replaceable cartridge), $l_z$ (displacement on the Z axis from robot flange to center axis of the instrument), $β$ (bending angle of the instrument). The coordinates of $P_1$ are given in Eq.4. The final coordinates of point E are given in Eq. 5. These coordinates are used to compute the position of the stapler with respect to the Tool Center Point (TCP) of the robot. As the stapler is mounted on the flange of the robot, the system allows the defining of the TCP to superpose over point E.

$$P_1 = (l_x, 0, l_z) \tag{4}$$

$$\begin{cases} X_E = l_x + l_y \cdot \cos(β) \\ Y_E = l_y \cdot \sin(β) \\ Z_E = l_z \end{cases} \tag{5}$$

## 6    Conclusions

This paper presents the development of a robotic circular stapler used for esophagogastric anastomosis. A SWOT analysis of the existent staplers is performed, and the obtained required characteristics are used for designing the robotic stapler. The surgical stapler is designed with an adapting flange, and it can be used with several serial and parallel robots. The design of the stapler is presented along with the torque required to actuate each mechanism of the instrument. The stapler is attached to a cognitive robot and the position of the robot during the procedure is presented along with the computation of the coordinates at the tip of the stapler in order to integrate the stapler with the robotic system. Future work will focus on developing the control system for the instrument, developing the prototype and performing functional validation tests using ex-vivo tissue. The limitation of this study consists in the overall dimensions of the instrument, that needs reduction in order to be able to be used in relevant experimental tests.

**Acknowledgements** This research was supported by the project "New smart and adaptive robotics solutions for personalized minimally invasive surgery in cancer treatment—ATHENA", funded by the European Union—NextGenerationEU and the Romanian Government, under the National Recovery and Resilience Plan for Romania, contract no. 760072/23 May 2023, code CF 16/15 November 2022, which in turn was through the Romanian Ministry of Research, Innovation and Digitalization, within Component 9, investment I8.

## 7    References

1. Qu, H.T., et al Esophageal cancer screening, early detection and treatment: Current insights and future directions. World Journal of Gastrointestinal Oncology 16(4), 1180-1191 (2024).
2. Mayo Clinic- "Esophageal Cancer-Diagnosis and Treatment: https://www.mayoclinic.org/diseases-conditions/esophageal-cancer/diagnosis-treatment/drc-20356090 , last accessed 2024/09/20.
3. American Cancer Society –"Treating Esophageal Cancer by Stage": https://www.cancer.org/cancer/types/esophagus-cancer/treating/by-stage.html, last accessed 2024.09.20.
4. Pisla, D.; Plitea, N.; Gherman, B.; Pisla, A.; Vaida, C. Kinematical analysis and design of a new surgical parallel robot. In Proceedings of the 5th International Workshop on Computational Kinematics, Duisburg, Germany, 6–8 May 2009; pp. 273–282.
5. Korpan, M., Puhr, H.C., Prager, G.W. et al. State-of-the-art therapy and innovative treatment strategies in esophageal squamous cell cancer. memo (2024). https://doi.org/10.1007/s12254-024-01006-3Journal 2(5), 99–110 (2016).
6. Fuccio L, et al: Brachytherapy for the palliation of dysphagia owing to esophageal cancer: A systematic review and meta-analysis of prospective studies. Radiother Oncol. 2017 Mar;122(3):332-339. doi: 10.1016/j.radonc.2016.12.034. Epub 2017 Jan 16. PMID: 28104297.

7. Tucan, P.; Vaida, C.; Horvath, D.; Caprariu, A.; Burz, A.; Gherman, B.; Iakab, S.; Pisla, D. Design and Experimental Setup of a Robotic Medical Instrument for Brachytherapy in Non-Resectable Liver Tumors. Cancers 2022, 14, 5841. https://doi.org/10.3390/cancers14235841.
8. Pisla, D.; Calin, V.; Birlescu, I.; Hajjar, N.A.; Gherman, B.; Radu, C.; Plitea, N. Risk Management for the Reliability of Robotic Assisted Treatment of Non-resectable Liver Tumors. Appl. Sci. 2020, 10, 52. https://doi.org/10.3390/app10010052.
9. Shaheen NJ, et al.: Radiofrequency ablation in Barrett's esophagus with dysplasia. N Engl J Med. 2009 May 28;360(22):2277-88. doi: 10.1056/NEJMoa0808145. PMID: 19474425.
10. Ketel, M. H., et al.: Nationwide Association of Surgical Performance of Minimally Invasive Esophagectomy With Patient Outcomes. JAMA network open, 7(4), e246556-e246556.
11. Saeidi, Hamed, et al. "A confidence-based shared control strategy for the smart tissue autonomous robot (STAR)." 2018 IEEE/RSJ International Conference on Intelligent Robots and Systems (IROS). IEEE, 2018.
12. Milone, M. et al. Fashioning esophagogastric anastomosis in robotic Ivor-Lewis esophagectomy: a multicenter experience. Langenbecks Arch Surg 409, 103 (2024). https://doi.org/10.1007/s00423-024-03290-3
13. Guerra, F., et al. Fully Robotic Side-to-Side Linear-Stapled Anastomosis During Robotic Ivor Lewis Esophagectomy. World J Surg 47, 2207–2212 (2023). https://doi.org/10.1007/s00268-023-07050-0
14. Wee, Jon O. et al. Early Experience of Robot-Assisted Esophagectomy With Circular End-to-End Stapled Anastomosis, The Annals of Thoracic Surgery, Volume 102, Issue 1, 253 – 259
15. Pisla, D. Vaida,C. Gherman, B., Chablat, D., Birlescu, I., Tucan, P. Innovative circular stapler for robotic assisted esophagogastric anastomosis, OSIM Patent pending 2024.
16. Tucan, P. Birlescu, I. Pusca, A. Gherman, B. Jucan, D. Antal, T. Vaida, C. Pisla, A. Chablat, D. Pisla, D. (2024). A Flexible Instrument for Robotic Assisted Minimally Invasive Esophagectomy. In: Rosati, G., Gasparetto, A., Ceccarelli, M. (eds) New Trends in Mechanism and Machine Science. EuCoMeS 2024. Mechanisms and Machine Science, vol 165. Springer, Cham. https://doi.org/10.1007/978-3-031-67295-8_8
17. Nottingham, J.M., McKeown, D.G.: Transhiatal esophagectomy, In: StatPearls Treasure Island (FL); 2024 Available from: www.ncbi.nlm.nih.gov/books/ NBK559196/, (2024).
18. Harrington C, Molena D.: Minimally invasive Ivor Lewis esophagectomy in 10 steps. JTCVS Tech. 2021 Aug 8;10:489-494, (2021).
19. Kono, K.: McKeown Esophagectomy. In: Lomanto, D., Chen, W.TL., Fuentes, M.B. Mastering Endo-Laparoscopic and Thoracoscopic Surgery. Springer, Singapore, (2023).
20. Barron, J. et al: Thoracoabdominal Esophagectomy: Then and Now, The Annals of Thoracic Surgery, In press, https://doi.org/10.1016/j.athoracsur.2023.12.017.
21. Servier Medical Art, https://smart.servier.com/citation-sharing/, last accessed 2024/09/20.
22. Panter Healthcare, http://www.pantherhealthcare.com/personal_protective_products2.html, last accessed 2024/09/20.
23. Shiraishi, O, et al. Circular Stapler Method for Avoiding Stricture of Cervical Esophagogastric Anastomosis. J Gastrointest Surg 26, 725–732 (2022). https://doi.org/10.1007/s11605-022-05266-4
24. Neura-Robotics, https://neura-robotics.com/products/maira, last accessed 2024/09/20.